\documentclass{article}
\usepackage[final]{nips_2017}
\usepackage[utf8]{inputenc} 
\usepackage[T1]{fontenc}    
\usepackage{hyperref}       
\usepackage{url}            
\usepackage{booktabs}       
\usepackage{amsfonts}       
\usepackage{nicefrac}       
\usepackage{microtype}      
\usepackage[T1]{fontenc}
\usepackage{amsmath,graphicx,subfigure}
\graphicspath{{Figures/}}
\usepackage{slashbox}

\newcommand{\specialcell}[2][c]{%
  \begin{tabular}[#1]{@{}c@{}}\vspace{-.05in}#2\end{tabular}}  
  
\title{Mosquito detection with low-cost smartphones: data acquisition for malaria research}

\author{
  Yunpeng Li\\
  Department of Engineering Science\\
  University of Oxford\\
  \texttt{yli@robots.ox.ac.uk}
  \And
  Davide Zilli\\
  Department of Engineering Science\\
  University of Oxford\\
  \& Mind Foundry Ltd.\\
  \texttt{dzilli@robots.ox.ac.uk}
  \AND
  Henry Chan\\
  Department of Engineering Science\\
  University of Oxford\\
  \texttt{tsunhenry@gmail.com}
  \And
  Ivan Kiskin\\
  Department of Engineering Science\\
  University of Oxford\\
  \texttt{ikiskin@robots.ox.ac.uk}
  \AND
  Marianne Sinka\\
  Department of Zoology\\
  University of Oxford\\
  \texttt{marianne.sinka@zoo.ox.ac.uk}
  \And
  Stephen Roberts\\
  Department of Engineering Science\\
  University of Oxford\\
  \& Mind Foundry Ltd.\\
  \texttt{sjrob@robots.ox.ac.uk}
  \AND
  Kathy Willis\\
  Department of Zoology\\
  University of Oxford\\
  \& Royal Botanic Gardens, Kew\\
  \texttt{kathy.willis@zoo.ox.ac.uk}
}

\begin{document}

\maketitle

\begin{abstract}
Mosquitoes are the only vector for malaria, causing
hundreds of thousands of deaths in the developing world each year.
Not only is the prevention of mosquito bites of paramount importance to the reduction of malaria transmission cases, but understanding in more forensic detail the interplay between malaria, mosquito vectors, vegetation, standing water and human populations is crucial to the deployment of more effective interventions.
Typically the presence and detection of mosquitoes is quantified through insect traps and human operations.
If we are to gather timely, large-scale data to improve this situation, we need to automate the process of mosquito detection and classification as much as possible.  In this paper,
we present a prototype mobile sensing system that acts as both a portable early warning device and an automatic acoustic data acquisition pipeline to help fuel scientific inquiry and policy.
The machine learning algorithm that powers the mobile system achieves excellent off-line multi-species detection performance while remaining computationally efficient. Further, we have conducted preliminary live mosquito detection tests using low-cost mobile phones and achieved promising results.
The deployment of this system for field usage in Southeast Asia and Africa is planned in the near future.
In order to accelerate processing of field recordings and labelling of collected data, we employ a citizen science platform in conjunction with automated methods, the former implemented using the
Zooniverse platform, allowing crowdsourcing on a grand scale.
\end{abstract}

\section{Introduction}

Malaria is one of the most severe public health problems in the developing world. The World Health Organization
estimated that there were 212 million
malaria cases worldwide in 2015,
leading to 429,000 related deaths~\cite{WHO2016}.
Vector-control efforts have achieved significant improvement in the past few decades~\cite{bhatt2015}.
However, the effect of malaria interventions remains poorly understood due to the absence of reliable surveillance data.

Malaria is transmitted through the bite of an infected \textit{Anopheles} mosquito.
Among the approximately 3,600 species of mosquitoes,
only about 60 out of the 450 Anopheles species transmit
malaria (i.e., are vectors)~\cite{neafsey2015,wilkerson2015}.
The ability to quickly detect the presence of these mosquito species is therefore crucial for control programmes and targeted intervention strategies.

As sensor-rich embedded devices, smartphones provide a perfect
platform for environmental sensing~\cite{lane2010}. 
They are programmable and equipped with
cheap yet powerful sensors, such as microphones, GPS, digital compasses and cameras.
The touch screen is ideal for displaying real-time feedback
to the user and inputting 
peripheral information in data acquisition. The built-in WiFi and cellular
access make them extremely useful for data streaming or synchronisation. 
These desirable properties have enabled a wide range of applications of smartphones
in environmental monitoring~
\cite{guo2015}.

The identification of disease-carrying mosquitoes by their flight tones has been researched 
for more than half a century~\cite{offenhauser1949,raman2007}.
However, to the best of our knowledge, there is no acoustic mosquito sensing pipeline
that is cost-effective and deployable in large-field studies. 
The HumBug project~\footnote{humbug.ac.uk} aims to accomplish
real-time acoustic mosquito detection using low-cost smartphones to alert users of the presence of vector species and guide control programmes, as well as relate such detections to geographic variables such as vegetation type and climate.
Following an initial proof-of-concept phase,
the system is going through field tests to improve
and evaluate the detection model with more field data.

In this paper, we describe our efforts in developing a real-time mosquito
detection app that acts both as an early warning device and an integral part of
an automatic data acquisition pipeline. Firstly, we present the machine learning algorithms deployed on low-cost smartphones for live detection; subsequently, we propose the
use of a citizen science platform to enable crowdsourcing
of data labels for improving the accuracy of the detection algorithm.

\section{Methods}

\subsection{Data acquisition}

\begin{figure}[htbp]
\centering
 \subfigure[Mozz Wear Interface.]{
 \includegraphics[scale=0.36]{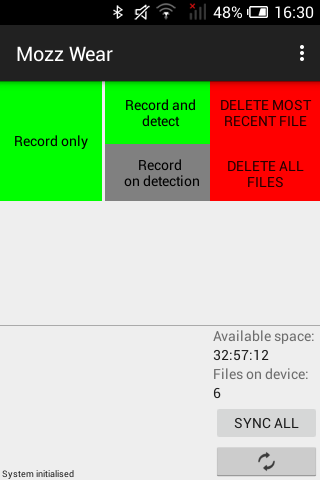}
 \label{fig:interface_main}
 }
 \subfigure[Mozz Wear interface (``Record and detection'' activated).]{
 \includegraphics[scale=0.36]{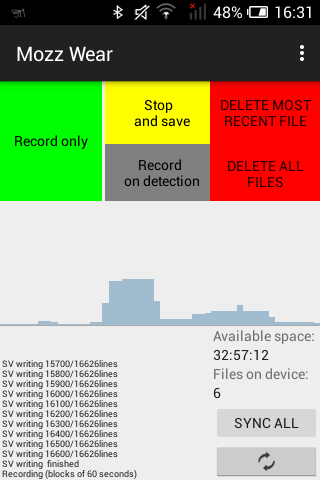}
 \label{fig:interface_detect}
 }
 \subfigure[Peripheral Information Input.]{
 \includegraphics[scale=0.36]{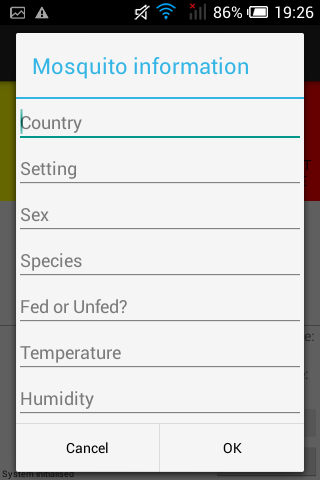}
 \label{fig:interface_category}
 }
 \caption{The Mozz Wear app.}
  \label{fig:app_interface}
\end{figure}

To enable acoustic detection of mosquitoes, an easy-to-use data
acquisition system needs to be set up for retrieving and transmitting data for
training of detection models and live prediction of mosquito presence.
The \emph{MozzWear} Android app we developed provides
a simple graphical user interface for sound recording and data synchronisation (Figure~\ref{fig:interface_main}). It currently supports the ``Record only'', the ``Record and detect'', and will support ``Record on detection'' functions. The ``Record and detect'' and the ``Record only'' both record sound.
In addition, the ``Record and detect'' function launches
the detection module and displays our real time predictions
of the mosquito presence (Figure~\ref{fig:interface_detect}).
Users of the app may have certain prior information to enter in the data collection stage, e.g. peripheral information of the environment, species
categories for cage mosquitoes.
The app provides a pull-down menu (Figure~\ref{fig:interface_category}) for users to enter such information (if available). All recordings
and corresponding peripheral information will be synchronised to
the online HumBug project database through WiFi or cellular connections.

\subsection{Classification pipeline}
\label{sec:classification}

The key objective of the classification algorithms for our specific task is high detection accuracy with a low computational and memory load. We describe our adopted feature extraction methods, two-stage multi-species classification algorithm, as well as the model training strategy.

\paragraph{Feature extraction:}
Audio information are often more distinctive in the frequency domain. Mel-frequency cepstral coefficients (MFCCs) are one of
the most widely used audio features in speech recognition
and acoustic scene classification~\cite{barchiesi2015}.
MFCCs provide a compact representations of spectral envelopes,
by performing the discrete cosine transform (DCT) on
the log power spectrum grouped according to the mel scale
of frequency bands. It usually compresses the high-dimensional
spectrum into a much lower space, e.g. 13-dimensional coefficients. Thus, it is well
suited for usage in low-power devices. In our experiments
comparing 11 different common audio features (omitted here), MFCCs lead to the best detection accuracy.

\paragraph{Two-stage multi-species classification:}
We adopt a two-stage classification paradigm for detection.
In the first stage, a binary class support vector machine (SVM)~\cite{cortes1995}
is used to detect the presence of mosquitoes. The SVM maps
the feature into a higher dimensional space for
more effective classification through the so-called kernel trick.

Once a mosquito is identified, a multi-species classification using the one-versus-one multi-class strategy with the SVM is used in the second stage to identify the exact mosquito species. We will soon introduce an option of
cost-sensitive SVMs in the first stage to provide a more direct control on the false positive rate~\cite{li2017d}.

\paragraph{Training strategy:}
\label{sec:training}
We explored different training strategies to achieve
the most effective classification
accuracy in testing, with a small amount of training data.
Only small amounts of training data are available, as labelling recordings is a manual and time-consuming process.

We divide recordings into audio clips with a duration of 0.1 seconds (which we call samples),
and assign labels to each clip according to data tagging results from humans.
A balanced training dataset is first created so that
each class has the same number of training samples.
The random sampling strategy, which randomly samples short audio clips without
replacement in the balanced training set, was found to produce the best detection performance (detailed results are again omitted).

\subsection{Data tagging and crowdsourcing}

\begin{figure}[htbp]
\centering
\includegraphics[scale=0.25]{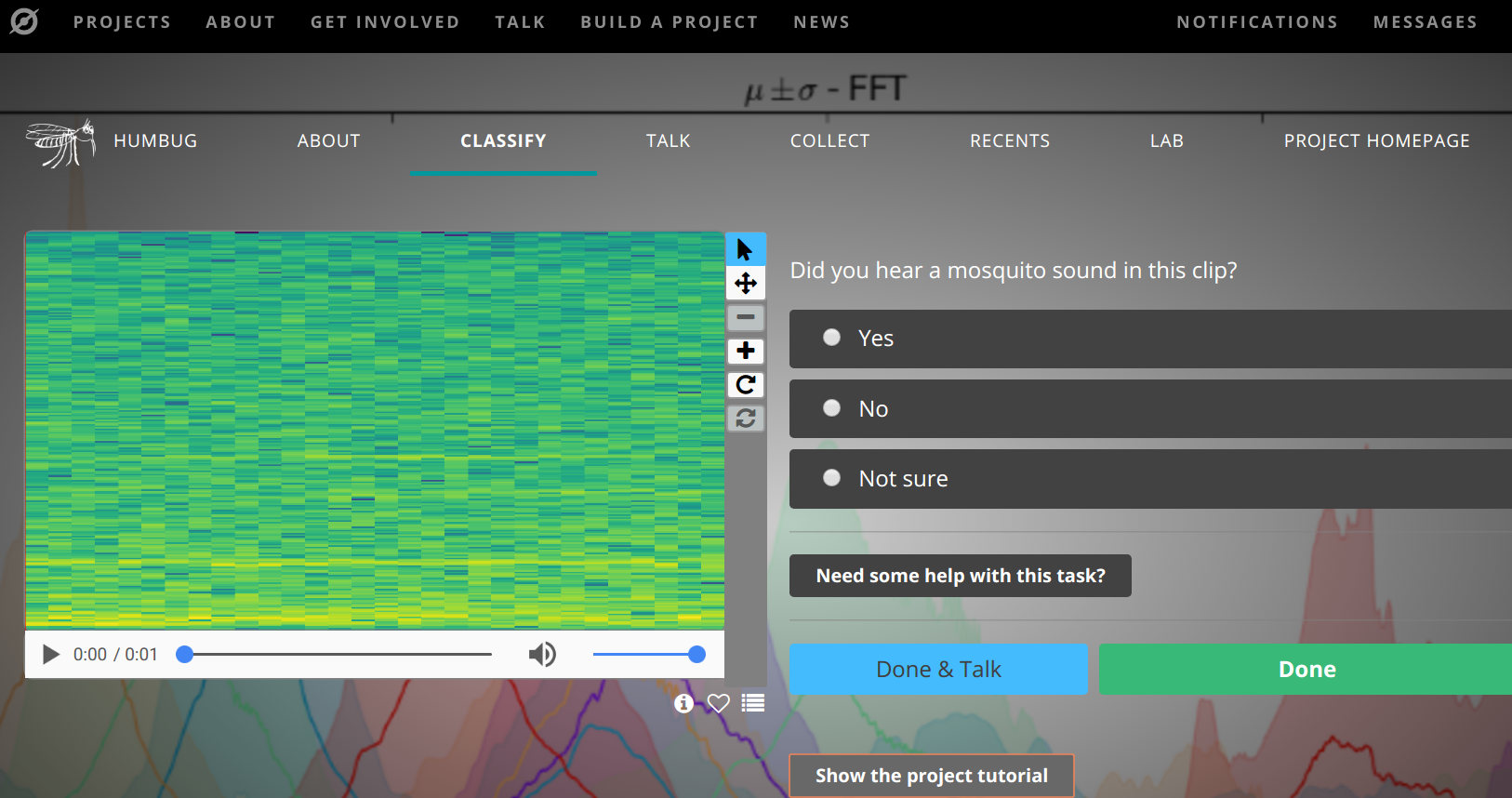}
\caption{\label{fig:zooniverse}
The classification page of the Zooniverse project.}
\end{figure}

Our current data labels were obtained through collective
data tagging from the project team members.
Data tagging involves marking segments of audio clips where mosquito sound can be heard.
With large-scale field deployment, the number of recordings
requiring data tagging will be beyond the capacity of
experts and researchers in the HumBug project. We hence resort to the power of crowdsourcing, creating a project
on Zooniverse\footnote{zooniverse.org},
the world's largest citizen science platform~\cite{simpson2014}, to solicit labels from millions of volunteers.
Volunteers listen to short sound clips and can see the
corresponding spectrograms, to give their decisions on whether
mosquito sound exists in audio clips (Figure~\ref{fig:zooniverse}).
Detection results described in Section~\ref{sec:classification} are used to filter data
uploaded to  Zooniverse. By only uploading
audio clips predicted
to contain mosquito sound, we may accelerate
the data tagging process.

\section{Experiments and Results}

We report off-line detection results
produced with a dataset containing audio recordings collected in the Centers for Disease Control and Prevention (CDC) in the US
and the US Army Military Research Unit in Kisumu, Kenya (USAMRU-K).
The CDC dataset contains sound recordings from 20 mosquito species. The amount of recordings for a majority species
is small -- for only 6 species we have no fewer than 62 0.1 second
samples. So, we include acoustic data from these 6 species
into the examined dataset. The USAMRU-K dataset contains sound recordings of the An. Gambiae species. 
We create a balanced dataset by downsampling 
data from each of these 7 mosquito species and background
recordings, so that each class contains 62 samples.
The balanced dataset simplifies training and leads to easier interpretation of prediction results on the test set.

100 trials were conducted by sampling
100 different copies of the balanced datasets through random initialisations. Hence data splitting as well as
algorithm random seeds are different
in different trials.
The ``random sampling'' training strategy described previously is adopted to train the detection algorithm with $50\%$ of samples in the balanced dataset. All tests were conducted with the \emph{MozzWear} installed on Alcatel One Touch 4009X mobile phones, available at a network operator's shop in the UK for \textsterling 20. The app is therefore suitable for low-end devices, aiding the ease of global deployment.


We report statistics of detection performance on test data among 100 trials in Table~\ref{tab:result}. Average detection accuracies
for different mosquito species vary from 0.68
to 0.92. The standard deviations (SD) calculated based on detection accuracies in 100 trials are low for most classes. $p$-values calculated based on the area-under-curve (AUC) of the macro-average of receiver operating characteristics (ROC) curves for each
class are less than $10^{-15}$ in
all trials.
Detection accuracies for \textit{Anopheles} mosquitoes, which are malaria vectors, are impressive albeit worse than detection accuracies for the other examined mosquito species. These results demonstrate that the current detection model, trained with a limited amount of data, has achieved effective multi-species detection with a dataset involving two experiments conducted in two different locations. Further field trips are planned to train more robust models and better assess detection performance using field data.

\begin{table}[htbp]
\centering
\tabcolsep=0.03cm
\caption{Statistics of detection accuracies with 7 mosquito
species. Results are generated from 100 trials.}
\label{tab:result}
\renewcommand{\arraystretch}{1.2}
\scalebox{0.9}
{
\begin{tabular}{|c|c|c|c|c|c|c|c|c|}
	\hline
	\backslashbox{Accuracy}{Class}
 & \specialcell{No\\mosquito} &
	\specialcell{\textit{Aedes}\\\textit{aegypti}} &
	\specialcell{\textit{Anopheles}\\\textit{quadrimaculatus}}& 
	\specialcell{\textit{Culex}\\\textit{tarsalis}} &
	\specialcell{\textit{Anopheles}\\\textit{albimanus}} &
	\specialcell{\textit{Culex}\\\textit{quinquefasciatus}} &
	\specialcell{\textit{Aedes}\\\textit{albopictus}} &
	\specialcell{\textit{Anopheles}\\\textit{Gambiae}} \\\hline
	Mean & 0.82 & 0.82 & 0.72 & 0.92 & 0.76
	 & 0.87 & 0.83 & 0.68\\\hline
	SD & 0.11 & 0.07 & 0.09 & 0.06 & 0.07
	 & 0.06 & 0.08 & 0.17\\\hline
	Min. & 0.5 & 0.58 & 0.41 & 0.74 & 0.57
	& 0.69 & 0.6 & 0.29 \\\hline
	Max. & 1.0 & 0.97 & 0.94 & 1.0 & 0.93 & 1.0
	& 1.0 & 1.0\\\hline
\end{tabular}}
\end{table}

\section{Conclusion}

Our acoustic mosquito detection system, despite using low-cost
smartphones, provides a promising avenue for live detection -- and species classification --
of mosquitoes known to vector malaria.
Our approach provides an automatic mosquito data acquisition pipeline with little additional cost. We demonstrate that the detection pipeline is efficient and can run smoothly on smartphones costing only
\$20. We are currently performing more field tests with this system.
Furthermore, our crowdsourcing platform provides an attractive solution
for large-scale data tagging from the growing database of acoustic recordings as we expand deployment.

\subsubsection*{Acknowledgments}

This work is part-funded by a Google Impact Challenge
award. The authors would like to thank
Paul I. Howell at the Centers for Disease Control and Prevention (CDC), BEI Resources in Atlanta, USA, 
Dustin Miller in CDC Foundation, Centers for Disease Control and Prevention in Atlanta, 
Dr. Sheila Ogoma, US Army Military Research Unit, Kisumu, Kenya (USAMRU-K), and Dr. Theeraphap Chareonviriyaphap, Kasersart University, Thailand for their collaborations on data collection and system deployment. 

\small
\bibliographystyle{apalike}
\bibliography{nips_2017_ML4D}

\end{document}